\documentclass{article}

\usepackage[final]{corl_2018} 
\usepackage{graphicx}
\usepackage{amsmath}
\usepackage{amssymb}
\usepackage{esvect}  
\usepackage{algorithmicx}
\usepackage{algorithm}
\usepackage{algpseudocode}
\usepackage{esvect}

\newcommand{\etal}{\emph{et al.}}

\title{Deep Object Pose Estimation for Semantic Robotic Grasping of Household Objects}

\author{
		\vspace{0.5ex}
    Jonathan Tremblay
    \hspace{4.5em} Thang To
    \hspace{2em} Balakumar Sundaralingam$^\dagger$ \\
		\vspace{0.5ex}
    \textbf{Yu Xiang}
    \hspace{4.5em} \textbf{Dieter Fox}
    \hspace{3em} \textbf{Stan Birchfield} \\
    NVIDIA\\
    \texttt{\{jtremblay,thangt,yxiang,dieterf,sbirchfield\}@nvidia.com} \\
		$^\dagger$Also with Univ. of Utah, \texttt{bala@cs.utah.edu}
}

\begin{document}
\maketitle

\vspace{-6mm}

\begin{abstract}

Using synthetic data for training deep neural networks for robotic manipulation holds the promise of an almost unlimited amount of pre-labeled training data, generated safely out of harm's way.  One of the key challenges of synthetic data, to date, has been to bridge the so-called \emph{reality gap}, so that networks trained on synthetic data operate correctly when exposed to real-world data.  We explore the reality gap in the context of 6-DoF pose estimation of known objects from a single RGB image.  We show that for this problem the reality gap can be successfully spanned by a simple combination of domain randomized and photorealistic data.  Using synthetic data generated in this manner, we introduce a one-shot deep neural network that is able to perform competitively against a state-of-the-art network trained on a combination of real and synthetic data.  To our knowledge, this is the first deep network trained only on synthetic data that is able to achieve state-of-the-art performance on 6-DoF object pose estimation.  Our network also generalizes better to novel environments including extreme lighting conditions, for which we show qualitative results.  Using this network we demonstrate a real-time system estimating object poses with sufficient accuracy for real-world semantic grasping of known household objects in clutter by a real robot.\footnote{Video and code: \url{https://research.nvidia.com/publication/2018-09_Deep-Object-Pose} .}
\end{abstract}




\section{Introduction}
	
\label{sec:intro}

In order for robots to operate safely and effectively alongside humans, they must be aware of their surroundings.  
One aspect of this awareness is knowledge of the 3D position and orientation of objects in the scene, often referred to as 6-DoF (degrees of freedom) pose. 
This knowledge is important to perform pick-and-place of objects, handoff from a person, or watch someone handle the object for imitation learning.
In this work we focus on rigid, known objects for which a prior training time to learn the appearance and shape of the objects is allowed.
Our goal is to infer the 3D pose of such objects, in clutter, from a single RGB image in real time for the purpose of enabling the robot to manipulate such objects.

While deep neural networks have been successfully applied to the problem of object detection in 2D \cite{dai2016nips:rfcn,redmon2017cvpr:yolo,Liu2016eccv:ssd}, they have only recently begun to be applied to 3D object detection and pose estimation \cite{rad2017iccv:bb8,xiang2018rss:posecnn,tekin2018cvpr:objpose}.  
Unlike 2D object detection, it is prohibitive to manually label data for 3D detection.
Due to this difficulty of collecting sufficiently large amounts of labeled training data, such approaches are typically trained on real data that are highly correlated with the test data (\emph{e.g.}, same camera, same object instances, similar lighting conditions).  As a result, one challenge of existing approaches is generalizing to test data that are significantly different from the training set.

Synthetic data is a promising alternative for training such deep neural networks, capable of generating an almost unlimited amount of pre-labeled training data with little effort.  
Synthetic data comes with its own problems, however.  Chief among these is the \emph{reality gap}, that is, the fact that networks trained on synthetic data usually do not perform well on real data without additional fine-tuning or other tricks.  A recently proposed solution to this problem is \emph{domain randomization} \cite{tobin2017iros:dr}, in which the training data is randomized in non-realistic ways so that, at test time, real data appears to the network as simply another variation.  Domain randomization has proved successful at detecting colored geometric shapes on a table \cite{tobin2017iros:dr}, flying a quadcopter indoors \cite{Sadeghi2017rss:cad2rl}, or learning visuomotor control for reaching \cite{Zhang2017arx:strdr} or pick-and-place \cite{James2017corl:etedr} of a brightly colored cube.

In this paper we explore a powerful complement to domain randomization (DR), namely, using photorealistic data \cite{tremblay2018arx:fat}.  We show that a simple combination of DR data with such photorealistic data yields sufficient variation and complexity to train a deep neural network that is then able to operate on real data without any fine-tuning.  Additionally, our synthetically-trained network generalizes well to a variety of real-world scenarios, including various backgrounds and extreme lighting conditions.  
Our contributions are thus as follows:
\begin{itemize}
  \item A one-shot, deep neural network-based system that infers, in near real time, the 3D poses of known objects in clutter from a single RGB image without requiring post-alignment.  This system uses a simple deep network architecture, trained entirely on simulated data, to infer the 2D image coordinates of projected 3D bounding boxes, followed by perspective-\emph{n}-point (P\emph{n}P) \cite{lepetit2009ijcv:epnp}.
  We call our system DOPE (for ``deep object pose estimation'').\footnote{Among the various meanings of the word \emph{dope}, we prefer to emphasize ``very good.''}  
  \item Demonstration that combining both non-photorealistic (domain randomized) and photorealistic synthetic data for training robust deep neural networks successfully bridges the reality gap for real-world applications, achieving performance comparable with state-of-the-art networks trained on real data.
	\item An integrated robotic system that shows the estimated poses are of sufficient accuracy to solve real-world tasks such as pick-and-place, object handoff, and path following.

\end{itemize}


\section{Approach}

We propose a two-step solution to address the problem of detecting and estimating the 6-DoF pose of all instances of a set of known household objects from a single RGB image.  First, a deep neural network estimates belief maps of 2D keypoints of all the objects in the image coordinate system.  Secondly, peaks from these belief maps are fed to a standard perspective-\emph{n}-point (P\emph{n}P) algorithm \cite{lepetit2009ijcv:epnp} to estimate the 6-DoF pose of each object instance.
In this section we describe these steps, along with our novel method of generating synthetic data for training the neural network.

\subsection{Network architecture}
Inspired by convolutional pose machines (CPMs) \cite{wei2016cvpr:cpm,cao2017cvpr:mppaf}, our one-shot fully convolutional deep neural network detects keypoints using a multistage
architecture. 
The feedforward network takes as input an RGB image of size $w \times h \times 3$ and branches to produce two different outputs, namely, belief maps and vector fields.  
There are nine belief maps, one for each of the projected 8 vertices of the 3D bounding boxes, and one for the centroids.
Similarly, there are eight vector fields indicating the direction from each of the 8 vertices to the corresponding centroid, similar to~\cite{xiang2018rss:posecnn}, to enable the detection of multiple instances of the same type of object.  
(In our experiments, $w=640$, $h=480$.)

The network operates in stages, with each stage taking into account not only the image features
but also the outputs of the immediately preceding stage.
Since all stages are convolutional, 
they leverage an increasingly larger effective receptive field as data pass through the network.  
This property enables the network to resolve ambiguities in the early stages due to small receptive fields by incorporating increasingly larger amounts of context in later stages.

Image features are computed by the first ten layers from VGG-19~\cite{simonyan2015iclr:vgg} (pretrained on ImageNet),
followed by two $3 \times 3$ convolution layers to reduce the feature dimension from 512 to 256, and from 256 to 128.
These 128-dimensional features are fed to the first stage consisting of three $3 \times 3 \times 128$ layers and one $1 \times 1 \times 512$ layer, followed by either a $1 \times 1 \times 9$ (belief maps) or a $1 \times 1 \times 16$ (vector fields) layer.  The remaining five stages are identical to the first stage, except that they receive a 153-dimensional input ($128 + 16 + 9 = 153$) and consist of five $7 \times 7 \times 128$ layers and one $1 \times 1 \times 128$ layer before the $1 \times 1 \times 9$ or $1 \times 1 \times 16$ layer.  All stages are of size $w/8$ and $h/8$, with ReLU activation functions interleaved throughout.  

\subsection{Detection and pose estimation}
\label{sec:detection}
After the network has processed an image, it is necessary to extract the individual objects from the belief maps.  In contrast to other approaches in which complex architectures or procedures are required to individuate the objects \cite{rad2017iccv:bb8,xiang2018rss:posecnn,kehl2017iccv:ssd6d,tekin2018cvpr:objpose}, our approach relies on a simple postprocessing step that searches for local peaks in the belief maps above a threshold, followed by a greedy assignment algorithm that associates projected vertices to detected centroids.  For each vertex, this latter step compares the vector field evaluated at the vertex with the direction from the vertex to each centroid, assigning the vertex to the closest centroid within some angular threshold of the vector.

Once the vertices of each object instance have been determined, a P\emph{n}P algorithm \cite{lepetit2009ijcv:epnp} 
is used to retrieve the pose of the object,
similar to \cite{tekin2018cvpr:objpose,rad2017iccv:bb8}. 
This step uses the detected projected vertices of the bounding box, the
camera intrinsics, and the object dimensions to recover the final
translation and rotation of the object with respect to the camera. 
All detected projected vertices are used, as long as at least the minimum number (four) are detected.

\subsection{Data generation}

A key question addressed by this research is how to generate effective training data for the network.  
In contrast to 2D object detection, for which labeled bounding boxes are relatively easy to annotate, 3D object detection requires labeled data that is almost impossible to generate manually.  Although it is possible to semi-automatically label data (using a tool such as LabelFusion \cite{marion2018icra:labelfusion}), the labor-intensive nature of the process nevertheless impedes the ability to generate training data with sufficient variation.  For example, we are not aware of any real-world training data for 6-DoF object pose estimation that includes extreme lighting conditions or poses.

To overcome these limitations of real data, we turn to synthetically generated data.  Specifically, we use a combination of non-photorealistic domain randomized (DR) data and photorealistic data to leverage the strengths of both.  As shown later in the experimental results, these two types of data complement one another, yielding results that are much better than those achieved by either alone.  
Synthetic data has an additional advantage in that it avoids overfitting to a particular dataset distribution, 
thus producing a network that is robust to lighting changes,
camera variations, and backgrounds. 

\begin{figure*}
\centering
\begin{tabular}{c|c}
domain randomized & photorealistic \\
\includegraphics[width=0.45\columnwidth]{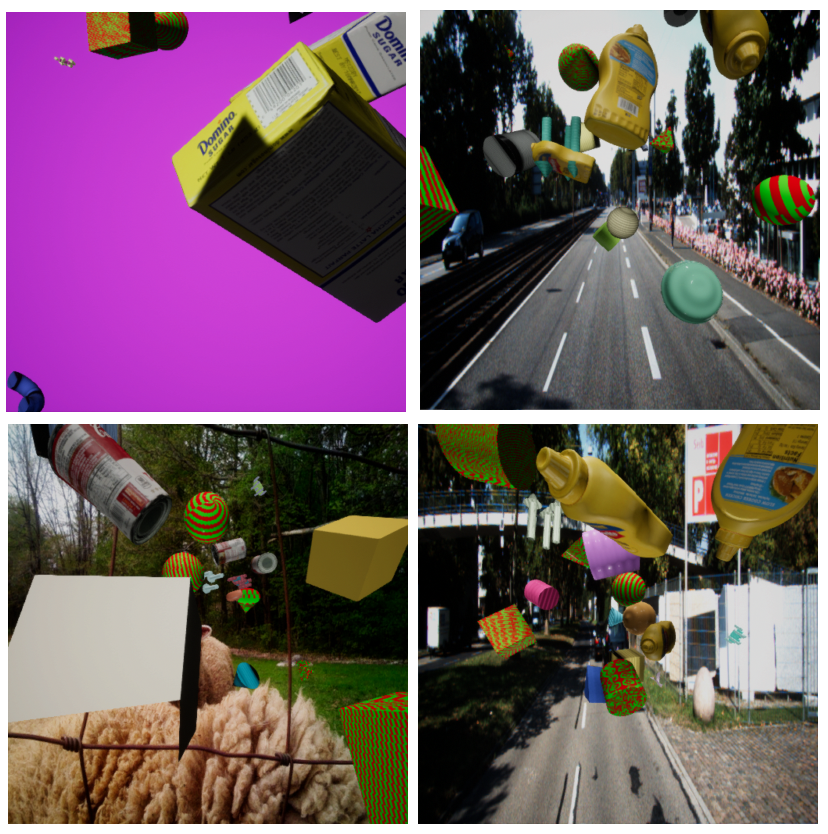}&
\includegraphics[width=0.45\columnwidth]{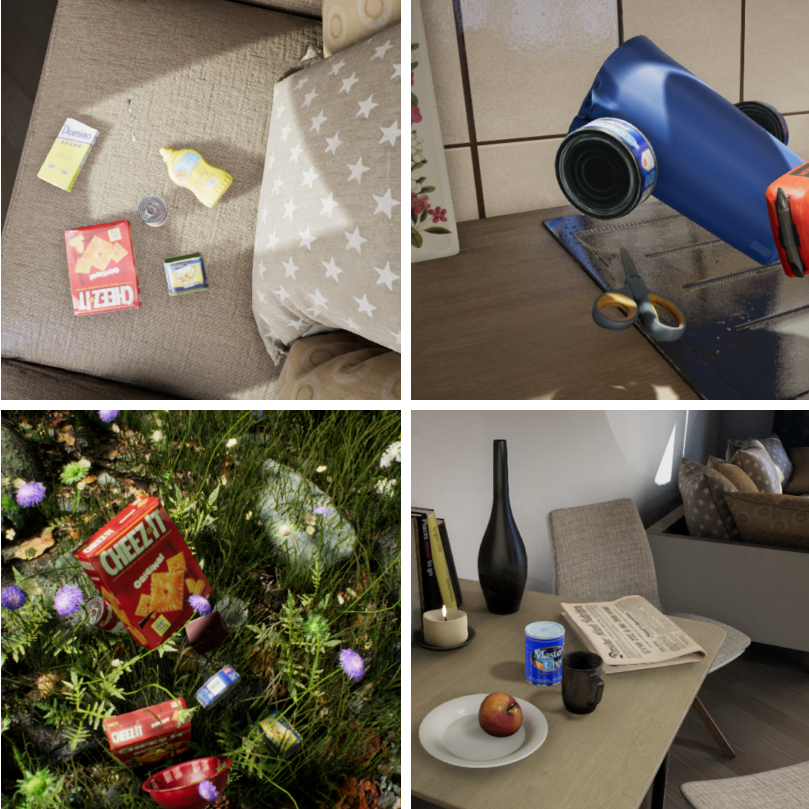}
\end{tabular}
\caption{Example images from our domain randomized (left) and photorealistic (right) datasets used for training.}
\label{fig:images}
\end{figure*}

Since our goal is robotic manipulation of the objects whose 6-DoF pose has been estimated, 
we leverage the popular YCB~objects~\cite{calli2015icar:ycb}.\footnote{\url{http://www.ycbbenchmarks.com/object-models}}  
These objects are readily available and suitable for manipulation by compliant, low-impedance human-friendly robots such as the Baxter.
Both the domain randomized and photorealistic data were created by placing the YCB object models in different virtual environments.
All data were generated by a custom plugin we developed for Unreal Engine 4 (UE4) called NDDS~\cite{to2018ndds}.
By leveraging asynchronous, multithreaded sequential frame grabbing, the plugin generates data 
at a rate of 50--100~Hz, which is significantly faster than either the default UE4 screenshot function or the publicly available Sim4CV tool~\cite{Mueller2018ijcv:Sim4CV}. 
Fig.~\ref{fig:images} shows examples of images generated using our plugin, illustrating the diversity and variety of both domain randomized and photorealistic data.

\vspace{2mm}\noindent \textbf{Domain randomization.}\hspace{1mm} 
The domain randomized images were created by placing the foreground objects
within virtual environments consisting of various distractor objects in front of a random background.  
Images were generated by randomly varying distractors, overlaid textures, backgrounds, object poses, lighting, and noise.
More specifically, the following aspects of the scene were randomized, similar to \cite{tremblay2018wad:car}:  number and types of distractors, selected from a set of 3D models (cones, pyramids, spheres, cylinders, partial toroids, arrows, etc.);
texture on the object of interest, as well as on distractors (solid, striped);
solid colored background, photograph (from a subset of 10,000 images from the COCO dataset \cite{Lin2014COCO}), or procedurally generated background image with random multicolored grid pattern;
3D pose of objects and distractors, sampled uniformly; 
directional lights with random orientation, color, and intensity; and
visibility of distractors.

\vspace{2mm}\noindent \textbf{Photorealistic images.}\hspace{1mm} 
The photorealistic data were generated by placing the foreground objects in 3D background scenes with physical constraints.  
Backgrounds were chosen from standard UE4 virtual environments such as a kitchen, sun temple, and forest.
These environments were chosen for their high-fidelity modeling and quality, as well as for the variety of indoor and outdoor scenes. 
For this dataset we included the same subset of 21 household YCB objects as in \cite{xiang2018rss:posecnn}.
Allowed to fall under the weight of gravity, and to collide with each other and with surfaces in the scene, these objects interact in physically plausible ways.
While the objects were falling, the virtual camera system was rapidly teleported to random azimuths, elevations, and distances with respect to a fixation point to collect data.
Azimuth ranged from --$120^\circ$ to +$120^\circ$ (to avoid collision with the wall, when present), elevation from $5^\circ$ to $85^\circ$, and distance from 0.5~m to 1.5~m.  
We recently made this dataset, known as Falling Things (FAT), publicly available~\cite{tremblay2018arx:fat}.


\section{Experimental Results}
\label{sec:result}

Since our goal is robotic manipulation of household objects, we focused our evaluation solely on images of the YCB objects, which incidentally is also the primary focus of the leading method for 6-DoF pose estimation, namely PoseCNN~\cite{xiang2018rss:posecnn}.  We compare our DOPE method to PoseCNN on both the YCB-Video dataset introduced in \cite{xiang2018rss:posecnn}, as well as on a dataset that we collected using a different camera, a variety of backgrounds, and extreme lighting conditions. Since PoseCNN has already been shown to 
outperform the single-shot method of Tekin \etal~\cite{tekin2018cvpr:objpose}, which itself beats the more complex networks of BB8~\cite{rad2017iccv:bb8} and SSD-6D~\cite{kehl2017iccv:ssd6d} on the standard LINEMOD~\cite{hinterstoisser2012accv:linemod} and Occluded-LINEMOD~~\cite{brachmann2014eccv:occlusion} datasets, the comparison of our method against PoseCNN is sufficient to answer the question as to whether our simple network architecture coupled with our novel synthetic data generation procedure are able to yield results that are on par with state-of-the-art.  Moreover, of the four algorithms just mentioned, PoseCNN is the only one with publicly available source code (at the time of this writing) capable of detecting YCB objects.\footnote{\url{https://github.com/yuxng/PoseCNN}}

\subsection{Datasets and metric}

For the first set of experiments we used the 
YCB-Video dataset~\cite{xiang2018rss:posecnn},
which is composed of $\sim$133k frames including pose
annotation for all objects in all frames.  The objects consist of a subset of 21 objects taken from the YCB dataset.  
We used the same testing set of 2,949 frames 
as suggested by Xiang \etal~\cite{xiang2018rss:posecnn}. 
We also collected our own dataset of four videos captured 
in a variety of extreme lighting conditions, by a Logitech~C960 camera.  In these videos we placed a subset of 5 objects from the 21 YCB objects in YCB-Video, including multiple instances of the same objects in many image frames.  To our knowledge, this is the first such dataset with multiple simultaneous instances.  The 5 objects are cracker box (003), sugar box (004), tomato soup can (005), mustard bottle (006), and potted meat can (010), where the numbers in parentheses indicate the object number in the YCB dataset.
These objects were selected for their easy availability at typical grocery stores, graspability, and visual texture.
We adopt the average distance (ADD) metric~\cite{hinterstoisser2012accv:linemod,xiang2018rss:posecnn} for evaluation, which is the average 3D Euclidean distance of all model points between ground truth pose and estimated pose. 


\subsection{Training}

For training we used $\sim$60k domain-randomized image frames mixed with $\sim$60k photorealistic image frames.  Whereas we used a separate set of $\sim$60k images in the former case (one dataset per object), in the latter case we used the same dataset for all objects, since it consists of the 21 YCB objects randomly interacting with one another.  
For data augmentation, Gaussian noise 
($\sigma = 2.0$), random contrast ($\sigma = 0.2$) and 
random brightness ($\sigma = 0.2$) were added.
For PoseCNN~\cite{xiang2018rss:posecnn}, we used the publicly available weights, which were trained using an undisclosed synthetic dataset and fine-tuned on images from separate videos of the YCB-Video dataset.

To avoid the vanishing gradients problem with our network, a loss was computed at 
the output of each stage, using the 
$L_2$ loss for the belief maps and vector fields. 
The ground truth belief maps were generated by placing 2D Gaussians 
at the vertex locations with $\sigma=2$~pixels. 
The ground truth vector fields were generated 
by setting pixels to the normalized $x$- and $y$-components of the vector pointing toward
the object's centroid.  
Only the pixels within a 3-pixel radius of each ground-truth vertex were set in this manner, with all other pixels
set to zero.
Wherever more than one vertex resided within this radius, one of them was selected at random to 
be used for generating these components.

Our network was implemented using PyTorch 
v0.4~\cite{paszke2017automatic:pytorch}.
The VGG-19 feature extractions were taken from 
publicly available trained weights in \texttt{torchvision} open models. 
The networks were trained for 60 epochs with a batchsize of 128. 
Adam~\cite{kingma2015ICLR:adam} was used as the optimizer with learning rate set at 0.0001.
The system was trained on an NVIDIA DGX workstation (containing 
4 NVIDIA P100 or V100 GPUs), 
and testing used an NVIDIA Titan~X.

\subsection{YCB-Video dataset}

\begin{figure*}
  \centering
    \includegraphics[width=.32\columnwidth,clip,
    ]
    {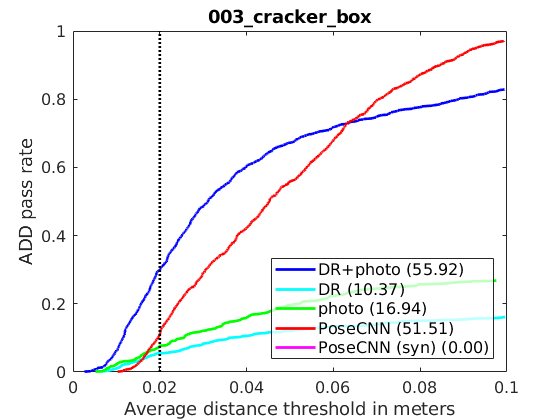}
    \includegraphics[width=.32\columnwidth,clip,
    ]
    {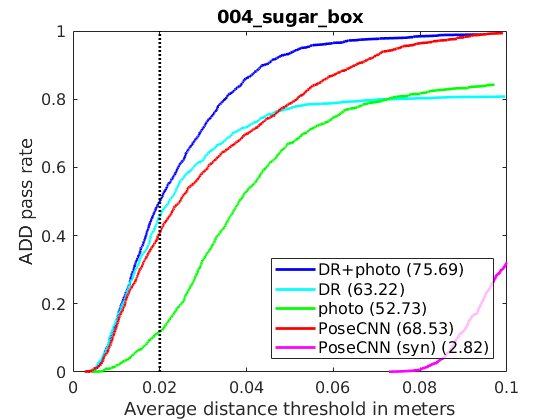}
    \includegraphics[width=.32\columnwidth,clip,
    ]
    {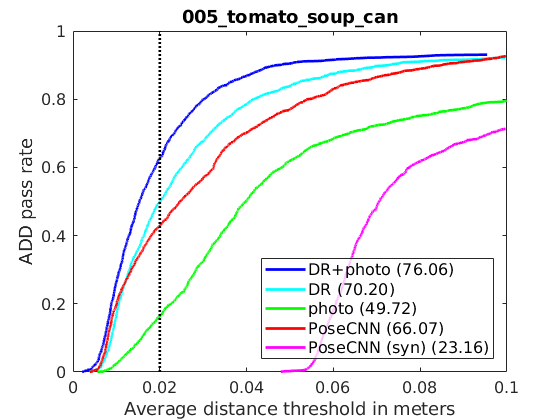}
    \includegraphics[width=.32\columnwidth,clip,
    ]
    {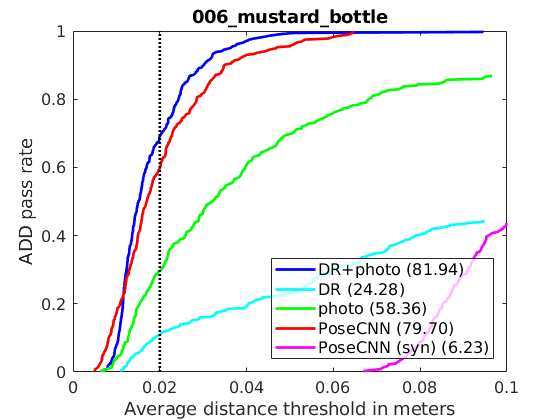}
    \includegraphics[width=.32\columnwidth,clip,
    ]
    {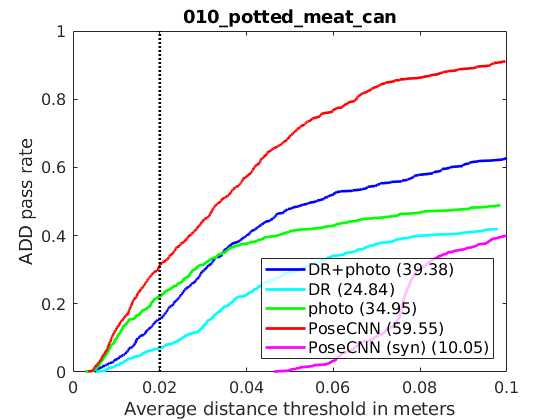}
    \caption{Accuracy-threshold curves for our DOPE method compared with PoseCNN~\cite{xiang2018rss:posecnn} for 5 YCB objects on the YCB-Video dataset.  Shown are versions of our method trained using domain-randomized data only (DR), synthetic photorealistic data only (photo), and both (DR+photo).  The numbers in the legend display the area under the curve (AUC).  The vertical dashed line indicates the threshold corresponding approximately to the level of accuracy necessary for grasping using our robotic manipulator (2~cm).  Our method (blue curve) yields the best results for 4 out of 5 objects.}
    \label{fig:results}
\end{figure*}

Fig.~\ref{fig:results} shows the accuracy-threshold curves
for five YCB objects
using the ADD metric for our DOPE method compared with PoseCNN.  
For our method we include three versions depending on the training set:  
using domain randomized data only ($\sim$60k synthetic images), 
using photorealistic data only  ($\sim$60k synthetic images), and using both  ($\sim$120k synthetic images).  
We also show results of PoseCNN trained on our DR+photo synthetic data, without real data.

These results reveal that our simple network architecture, trained only on synthetic data, is able to achieve results on par with the state-of-the-art PoseCNN method trained on a mixture of synthetic and real data.  These results are made more significant when it is noted that the real images on which PoseCNN was trained were taken from other videos in the same YCB-Video dataset, meaning that they were captured by the same camera under similar conditions as the test data.  Even so, our synthetic-only approach achieves higher area under the curve (AUC) for 4 of the 5 objects, as well as better results at thresholds less than 2~cm, which is about the limit of graspability error for the Baxter robot's parallel jaw end effector.  The reason our network fails to detect many of the potted meat can instances is due to severely occluded frames in which only the top of the can is visible; since our synthetic data does not properly model the highly reflective metallic material of the top surface, the network does not recognize these pixels as belonging to the can.  
We leave the incorporation of such material properties to future work.

\subsection{Extreme lighting dataset}

\begin{figure*}
  \centering
  \begin{tabular}{ccccc}
		\raisebox{4ex}{\rotatebox{90}{PoseCNN \cite{xiang2018rss:posecnn}}} & 	
    \includegraphics[width=.21\columnwidth,clip,
    trim={100px 100px 150px 0px}]
      {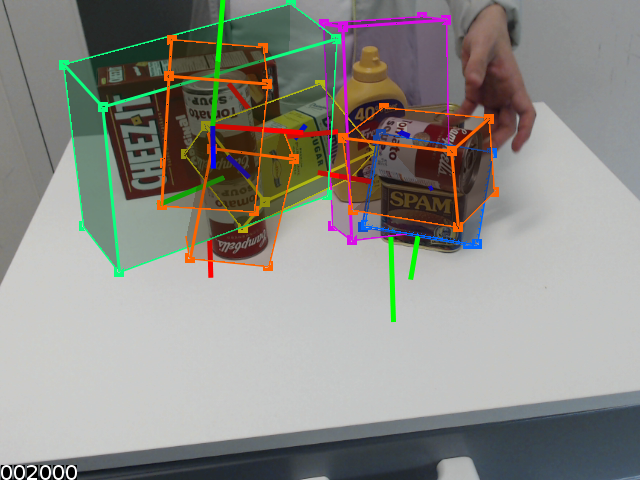} &
    \includegraphics[width=.21\columnwidth,clip,
    trim={150px 50px 100px 50px}]
      {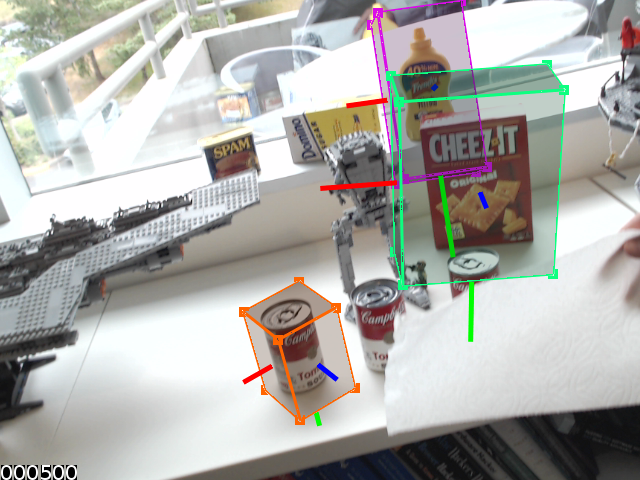} &
    \includegraphics[width=.21\columnwidth,clip,
    trim={100px 100px 150px 0px}]
      {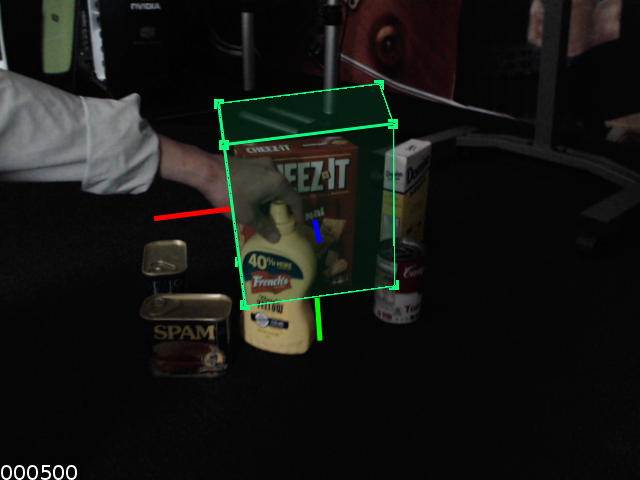} &
    \includegraphics[width=.21\columnwidth,clip,
    trim={50px 10px 200px 90px}]
      {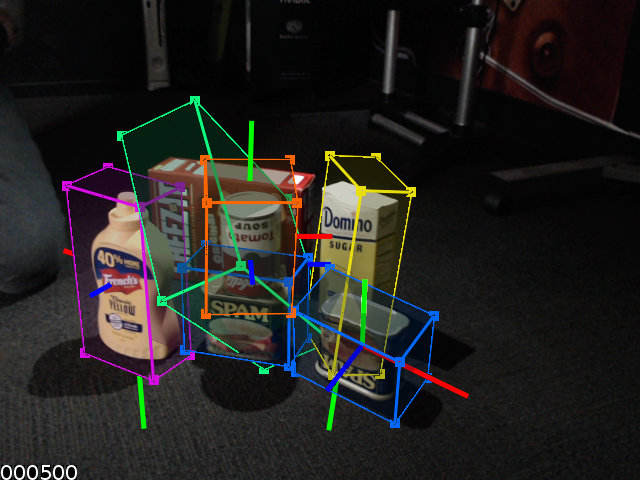} \\
      \vspace{-4mm}\\  
      \hline\\
      \vspace{-6.5mm}\\  
    \raisebox{4ex}{\rotatebox{90}{DOPE (ours)}} & 
    \includegraphics[width=.21\columnwidth,clip,
    trim={100px 100px 150px 0px}]
      {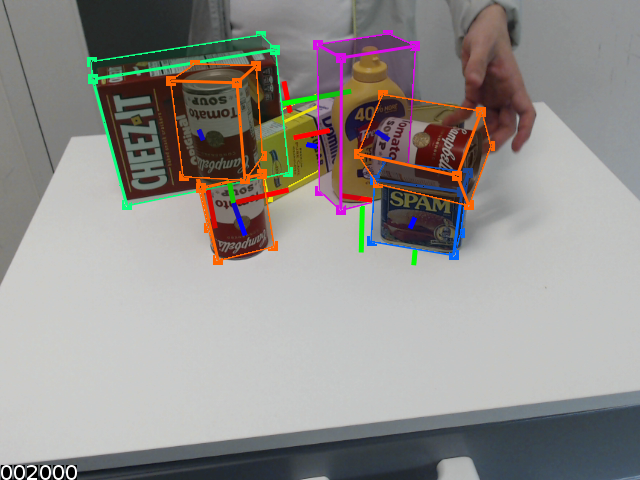} &
    \includegraphics[width=.21\columnwidth,clip,
    trim={150px 50px 100px 50px}]
      {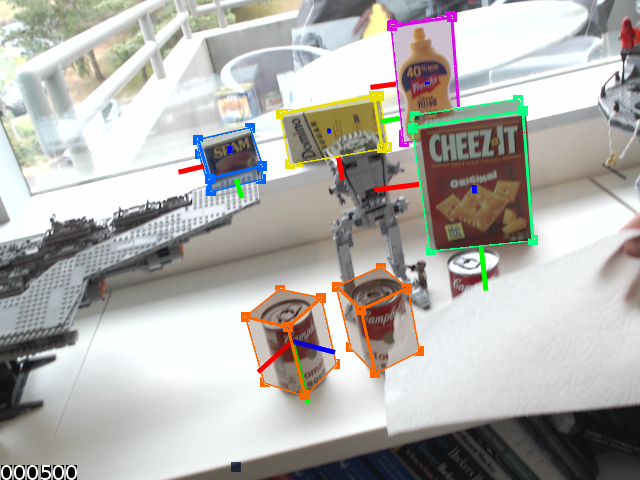} &
    \includegraphics[width=.21\columnwidth,clip,
    trim={100px 100px 150px 0px}]
      {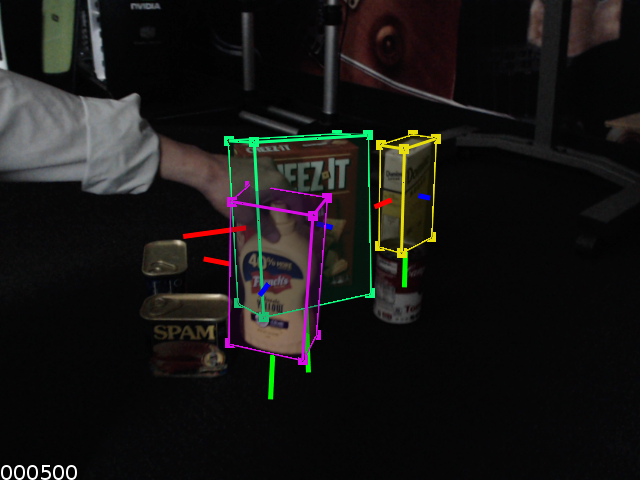} &
    \includegraphics[width=.21\columnwidth,clip,
    trim={50px 10px 200px 90px}]
      {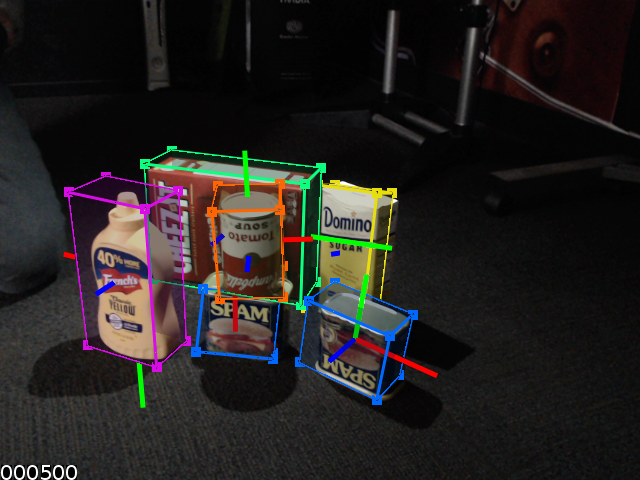}
    \end{tabular}
  \caption{Pose estimation of YCB objects on data showing extreme lighting conditions.  {\sc Top:} PoseCNN \cite{xiang2018rss:posecnn}, which was trained on a mixture of synthetic data and real data from the YCB-Video dataset \cite{xiang2018rss:posecnn}, struggles to generalize to this scenario captured with a different camera, extreme poses, severe occlusion, and extreme lighting changes.   
	{\sc Bottom:}  Our proposed DOPE method generalizes to these extreme real-world conditions even though it was trained only on synthetic data; all objects are detected except the severely occluded soup can (2nd column) and three dark cans (3rd column). }
  \label{fig:results_extreme}
\end{figure*}

To test the generalization performance of our DOPE approach compared with PoseCNN, we ran the two networks on the extreme lighting videos that we collected.  
Fig.~\ref{fig:results_extreme} shows four images from these videos,\footnote{Results on other video frames are similar to those shown; see the accompanying video.} showing qualitatively that 
the poses found by DOPE trained using our combined synthetic data generation procedure are more robust and precise than those of PoseCNN.
These qualitative results highlight the promise of using synthetic data to yield better generalization across cameras and lighting conditions than using real data (due to the difficulty of labeling the latter).

\subsection{Additional experiments}

To test the effect of dataset size, we trained our network on DR data only using 2k, 10k, 20k, 50k, 100k, 200k, 300k, $\ldots$, 1M.  
(We arbitrarily chose the sugar box for this experiment.) 
The biggest performance increase 
occurred from 2k (0.25 AUC) to 10k (54.22 AUC), results
saturated around 100k (60.6 AUC), and 
the highest value was achieved with 300k (66.64 AUC).
This experiment was then repeated using
photorealistic images instead, with
the highest value achieved at 600k (62.94 AUC).
Both of these results are far below the DR + photo result (77.00 AUC) reported above with 120k.
These results confirm that our proposed method of mixing DR and photorealistic 
synthetic data achieves greater success at domain transfer 
than either DR or photorealistic images alone. 


We also compared different mixing percentages.
That is, we trained with increments of 
10\% of each dataset while keeping the other at 100\% ({\em e.g.}, 
using 50\% of the DR dataset with 100\% of the photorealistic dataset). 
The performance was comparable for all networks as long as 
at least 40\% of either dataset was included. 

\begin{figure*}
  \centering
  \begin{tabular}{cc}
    \includegraphics[width=.4\columnwidth,clip]
    {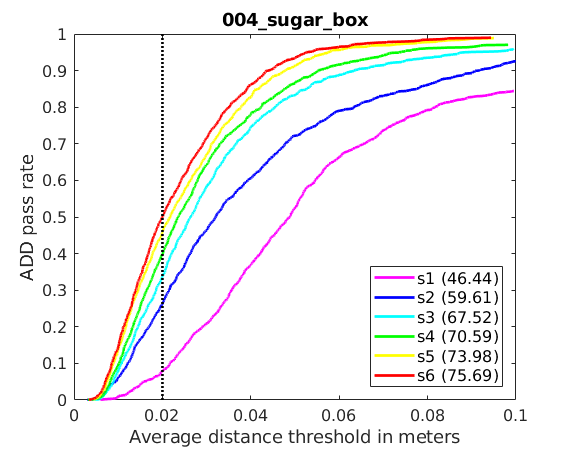} &
    \raisebox{20mm}{
    \begin{small}
    \begin{tabular}{l|cc}
      & speed (ms) & AUC \\
      \hline
      1 stage  & \, 57 & 46.44\\
      2 stages & \, 88 & 59.61\\
      3 stages & 124 & 67.52\\
      4 stages & 165 & 70.59\\
      5 stages & 202 & 73.98\\
      6 stages & 232 & 75.69\\
    \end{tabular}  
    \end{small}
    }
  \end{tabular}
  \caption{Accuracy-threshold curves with various numbers of stages, showing the benefit of additional stages to resolve ambiguity from earlier stages. 
  The table shows the total execution time, including object extraction and P\emph{n}P, and performance of the system for different numbers of stages.}
  \label{fig:performance}
\end{figure*}

Fig.~\ref{fig:performance} shows the impact of the number of stages on both accuracy and computing time.  
As expected, additional stages yield higher accuracy at the cost of greater computation.




\subsection{Robotic manipulation}

For our purposes, the ultimate test of a pose estimation method is whether its accuracy is sufficient for robotic grasping.  We attached a Logitech~C960 camera to the waist of a Baxter robot, and calibrated the camera to the robot base using a standard checkerboard target visible to both the Logitech and wrist cameras.  The parallel jaw gripper moves from an opening of approximately 10~cm to 6~cm, or from 8~cm to 4~cm, depending on how the fingers are attached.  Either way the travel distance of the gripper is just 4~cm, meaning that the error can be no more than 2~cm on either side of the object during the grasp.  

For quantitative results, we placed the 5 objects, amidst clutter, at 4 different locations on a table in front of the robot, at 3 different orientations for each of the 4 locations.  The robot was instructed to move to a pregrasp point above the object, then execute a top-down grasp, yielding 12 trials per object.  Of these 12 attempts, the number of successful grasps were as follows:  10 (cracker), 10 (meat), 11 (mustard), 11 (sugar), and 7 (soup).  The difficulty of the soup arises because the round geometry is unforgiving to top-down grasps.  When we repeated the experiment with the soup lying on its side, this number increased to 9.  These experiments demonstrate that our DOPE pose estimation method is robust enough in real-world conditions to execute successful grasps, without any closed-loop servoing, in a majority of cases.  Sources of error include the pose estimation algorithm, mis-calibration between the camera and robot, and imprecise robot control.  Although it is difficult to pinpoint the exact cause of error in each case, anecdotally the imprecise control of the robot appears to play a disproportionate role in many cases we observed.  For better control, a method like that of Levi~\etal~\cite{levi2015ichr:mpcis} might be appropriate.

\begin{figure}
  \begin{tabular}{c|c|c}
    pick & move & place \\
    \includegraphics[width=.3\columnwidth]
      {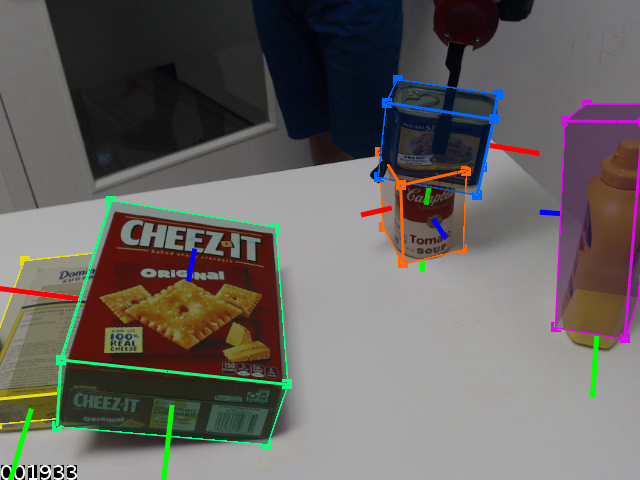} &
    \includegraphics[width=.3\columnwidth]
      {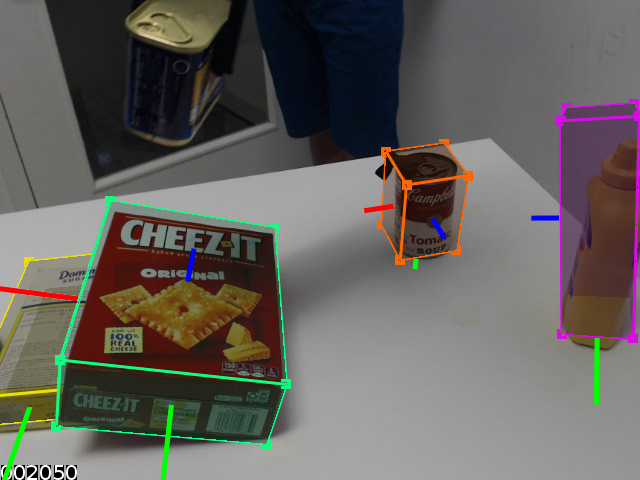} &
    \includegraphics[width=.3\columnwidth]
      {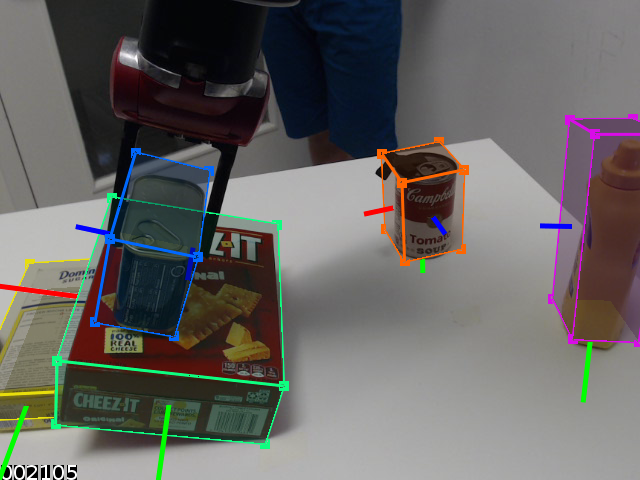} 
  \end{tabular} 
  \caption{Robotic pick-and-place of a potted meat can on a cracker box.  Note that the can is initially resting on another object rather than on the table, and that the destination box is not required to be aligned with the table, since the system estimates full 6-DoF pose of all objects.  Note also that the can is aligned with the box (as desired) and within a couple centimeters of the center of the box.}

  \label{fig:pick_place}
\end{figure}

To our knowledge, these results are the first repeatable experiments of semantic robotic grasping of household objects using the output of a real-time 6-DoF object pose estimator.  Unlike the semantic grasping work of \cite{jang2017corl:semgrasp}, our system is not limited to top-down grasps of static objects.  Rather, because it estimates 6-DoF pose in real time, our system is capable of performing pick-and-place operations where one object is placed on top of another object, as shown in Fig.~\ref{fig:pick_place}, as well as grasping the object out of a human's hand (\emph{i.e.}, handoff from the human to the robot), and causing the object to follow the 6-DoF path of another object in real time.  Unlike the approach of Tobin~\etal~\cite{tobin2017iros:dr}, objects are not restricted to be on the table when grasped but can be anywhere in the visible workspace.

%




\section{Related Work}


\textbf{Object detection and 6-DoF pose estimation.}
The problem of object detection and 6-DoF pose estimation from RGB images has been addressed by many researchers using traditional computer vision methods, such as 
\cite{hinterstoisser2012pami:linemod,pauwels2015iros:simtrack,pauwels2016csvt:hundreds,Tjaden2017:pose}.  
Recently, several deep learning-based approaches have appeared that operate directly on RGB images~\cite{rad2017iccv:bb8,xiang2018rss:posecnn,kehl2017iccv:ssd6d,tekin2018cvpr:objpose}, showing improved performance with respect to difficulties such as occlusion. 
Of these, BB8~\cite{rad2017iccv:bb8} and PoseCNN~\cite{xiang2018rss:posecnn} first segment the objects, then estimate the object pose, whereas the approaches of Tekin \etal~\cite{tekin2018cvpr:objpose}
and Kehl \etal~\cite{kehl2017iccv:ssd6d} rely on the single-shot networks of YOLO and SSD, respectively.  
All of these networks were trained on real data, thus limiting their ability to generalize to new scenarios (\emph{e.g.}, lighting conditions).
Moreover, directly regressing to the pose bakes the camera intrinsics directly into the network weights, causing additional errors when applying the network to a different camera.
Our approach differs from these methods in the following ways:  1) We train only on synthetic data, 2) our single-shot belief map network architecture is simpler, and 3) we demonstrate the accuracy of the system in the context of real-world robotic manipulation.
Moreover, our approach does not rely on post-refinement of the estimated pose, and our use of P\emph{n}P allows the method to operate on different camera intrinsics without retraining.  
For alternate approaches to related problems, see \cite{Mousavian2017cvpr:3dbb,pavlakos2017icra:6dof,mitash2017iros:self,Sundermeyer2018eccv:implicit,Suwajanakorn2018nips:keypoints}.

%
%
%
%


\textbf{Synthetic data for training.}  Given the deep learning hunger for large amounts of labeled training data, a recent research trend 
has focused on providing synthetic datasets for training 
\cite{butler2012eccv,handa2015arx:sn,dosovitskiy2015iccv:flownet,mayer2016cvpr:flythings,qiu2016arx:uncv,zhang2016arx:unst, mccormac2017iccv:snrgbd,ros2016cvpr:syn,richter2016eccv:pfd,gaidon2016cvpr:vkitti,Mueller2018ijcv:Sim4CV,tsirikoglou2017arx}.
Most of these datasets are photorealistic, thus requiring 
significant modeling effort from skilled 3D artists. 
To address this limitation, 
domain randomization~\cite{tobin2017iros:dr,Sadeghi2017rss:cad2rl} has been proposed as an inexpensive alternative that forces the network 
to learn to focus on essential features of the image by randomizing the training input in non-realistic ways. 
While domain randomization has yielded promising results on several tasks, it has not yet been shown to produce state-of-the-art results compared with real-world data.
In \cite{tobin2017iros:dr,tremblay2018wad:car}, for example, the authors found that fine-tuning with real data was necessary for domain randomization to compete with real data.
We hypothesize that domain randomization alone is not sufficient for the network to fully understand a scene, given its non-realism and lack of context.
Thus, our approach of using photorealistic data to complement domain randomization can be seen as 
a promising solution to this problem.


\section{Conclusion}
\label{sec:conclusion}

We have presented a system for detecting 
and estimating the 6-DoF pose of known objects using a novel architecture and 
data generation pipeline. 
The network employs multiple stages to refine ambiguous 
estimates of the 2D locations of projected vertices of each object's 3D bounding cuboid.
These points are then used to predict the final pose using P\emph{n}P, assuming known camera intrinsics and object 
dimensions. 
We have shown that a network trained only on synthetic data can achieve state-of-the-art performance
compared with a network trained on real data,
and that the resulting poses are of sufficient accuracy for robotic manipulation.
Further research should be aimed at increasing the number of objects, handling symmetry, and incorporating closed-loop refinement to increase grasp success.


\clearpage


\bibliography{pose}  

\end{document}